\documentclass{svproc}
\usepackage{url}

\usepackage[authoryear,round]{natbib}  
\usepackage{amsmath}
\usepackage{amsfonts}
\usepackage{bm}
\usepackage{algorithm}
\usepackage{algorithmic}
\usepackage{xspace}
\usepackage{enumitem}
\usepackage{multirow}
\usepackage{rotating}
\usepackage{subfigure}

\newcommand{\ie}{\emph{i.e.,}\xspace}

\newcommand{\name}{GCD\xspace}
\newcommand{\etal}{\emph{et al.}\xspace}
\newcommand{\etc}{\emph{etc.}\xspace}
\newcommand{\ignore}[1]{}

\newtheorem{myDef}{Definition}

\begin{document}
\mainmatter              
\title{Are $L_2$ adversarial examples intrinsically different?}
\titlerunning{L2 adversarial examples property}  
%
\author{Mingxuan Li\inst{1} \and Jingyuan Wang\inst{2} \and Yufan Wu\inst{2}
}
%
%
%
\institute{Brown University,\\
\email{mingxuan\_li@brown.edu},\\
\and
Beihang University,\\
\email{\{jywang, wuyufan\}@buaa.edu.cn}}

\maketitle              

\begin{abstract}
Deep Neural Network (DDN) has achieved notable success in various tasks, including many security concerned scenarios. However, a considerable amount of work has proved its vulnerability to adversaries. We unravel the properties that can intrinsically differentiate adversarial examples and normal inputs through theoretical analysis. That is, adversarial examples generated by $L_2$ attacks usually have larger input sensitivity which can be used to identify them efficiently. We also found that those generated by $L_\infty$ attacks will be different enough in the pixel domain to be detected empirically. To verify our analysis, we proposed a \textbf{G}uided \textbf{C}omplementary \textbf{D}efense module (\textbf{GCD}) integrating detection and recovery processes. When compared with adversarial detection methods, our detector achieves a detection AUC of over 0.98 against most of the attacks. When comparing our guided rectifier with commonly used adversarial training methods and other rectification methods, our rectifier outperforms them by a large margin. We achieve a recovered classification accuracy of up to 99\% on MNIST, 89\% on CIFAR-10 and 87\% on ImageNet subsets against $L_2$ attacks. Furthermore, under the white-box setting, our holistic defensive module shows a promising degree of robustness. Thus, we confirm that at least $L_2$ adversarial examples are intrinsically different enough from normal inputs both theoretically and empirically. And we shed lights upon designing simple yet effective defensive methods with these properties.
\end{abstract}

\section{Introduction}

Deep Learning has achieved great success, especially in the field of computer vision. Multiple breakthroughs make it possible to deploy these models into real life applications, including autonomous driving~\citep{lin2018architectural}, face recognition~\citep{parkhi2015deep}, industrial automation~\citep{meyes2017motion} and \etc. 
However, firstly pointed out by Szegedy et al.~\citep{2013szegedy}, deep neural networks can be fooled by adding small perturbations to the inputs, which are imperceptible to human eyes. Since then, lots of work have proved the weakness of deep networks, particularly under computer vision context~\citep{akhtar2018threat}. Such weakness draws attention of academia to the security aspect of deep neural networks.

Since then, many defense methods have been proposed, including robust optimization~\citep{madry2017pgd,fgsm}, model architectures modification~\citep{BuckmanRRG18, DhillonALBKKA18, SamangoueiKC18, DBLP:conf/iclr/XieWZRY18, DBLP:conf/iclr/SongKNEK18}, input reconstruction~\citep{GuoRCM18} and some detection based methods~\citep{DBLP:conf/ccs/MengC17, DBLP:conf/iclr/MetzenGFB17, DBLP:conf/iclr/HendrycksG17a,DBLP:journals/corr/BhagojiCM17,DBLP:conf/iccv/LiL17}. Though making seemingly promising progress against various emerging strong attacks, these methods still have some distance to alleviating real world security concerns. These steps in front include requiring large volume of computational resources~\citep{madry2017pgd}, being unresistant to white-box attacks~\citep{carlini2017adversarial}, non-scalable nor tested in bigger datasets and sometimes in need of modifying protected model architectures.

Inspired by the robustness analysis of a single layer neural network, we answer the question whether adversarial examples are intrinsically different from clean inputs or not. It turns out that adversarial examples generated by $L_2$ which we can leverage when designing detection methods. Additionally, we found empirically that $L_\infty$ attacks have different properties in pixel domains. 

\begin{figure}[t!]
    \centering
    \includegraphics[width = 0.75\textwidth]{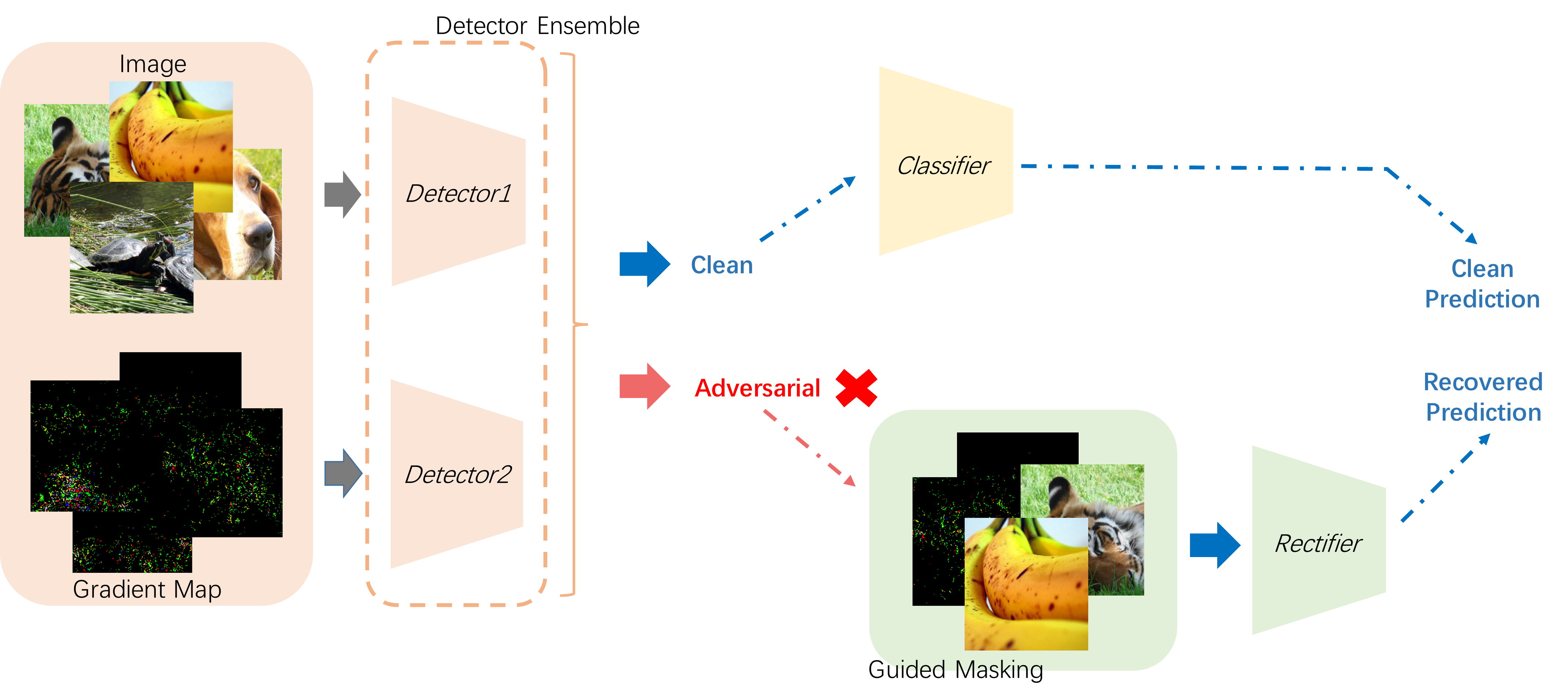}
    \caption{The framework of $\name$. Information entropy based detectors ensemble takes image and gradient sensitivity of base model. If the input image is identified as adversarial, a further rectification process is required. It will be patched according to gradient sensitivity of detectors. The final prediction of $\name$ only comes from the base model being protected when detector labels input as clean.}
    \label{pic:full arch}
\end{figure}

Thus, we proposed to build a defensive module which makes use of both properties in a complementary way, called \textbf{G}uided \textbf{C}omplementary \textbf{D}efense module, namely, \textbf{\name}. The \name model consists of a detection and rectification pipeline. The detection process captures adversarial examples using a gradient-domain image sensitivity monitor and a pixel-domain image sensitivity monitor. The rectification process recovers true predictions of perturbed images using a input sensitivity-guided rectifier (see Fig.~\ref{pic:full arch}). 

The experiment results over three commonly used datasets show that our model can completely defeat several state-of-the-art attacking algorithms and increase some attacks' cost significantly. Though there are some preliminary work on this topic~\citep{dhaliwal2018gradient, 2018qinghua_imagegrad}, we are the first to show what makes adversarial examples different theoretically and utilize those properties to build effective defense module in a principled way.
The contribution of this paper can be concluded as:
\begin{itemize}[topsep=0pt, partopsep=2pt, itemsep=0pt]    
\item We find the intrinsic property \textbf{analytically} that can differentiate between $L_2$ adversarial and benign examples. To our best knowledge, we are the first to justify the usage of gradient based detector against $L_2$ adversarial examples theoretically. 

\item The information entropy based detector ensemble in our proposed $\name$ module achieves a graceful performance (about 0.98 detection AUC) on MNIST, CIFAR-10 and ImageNet under attacks of state-of-the-art algorithms, including Deepfool~\citep{moosavi2016deepfool}, CW~\citep{carlini2017cw}, DDN(champion of NIPS Adversarial Vision Challenge 2018)
~\citep{rony2019ddn}, and PGD~\citep{madry2017pgd}.

\item We propose a rectification process guided by the gradient information from detectors to recover the true label of inputs. After rectification, we achieve a recovered classification accuracy of 99\% on MNIST, 87\% on CIFAR10, 83\% on ImageNet subset on average, surpassing adversarial trained model and other rectification methods by a large margin.

\item $\name$ does not require extra modification to the base model being protected. This feature makes it suitable for neural network systems already deployed.

\end{itemize}


\section{The Basic Idea}~\label{sec:motivation}
In this section, we will briefly introduce our motivation and the inherent idea of our method.
\begin{myDef}[Classifier]
    We define a neural network classifier as a function 
    $f: \mathbb{R}^M \rightarrow [0,1]^L$ 
    with a softmax function as its output.
\end{myDef}

\begin{myDef}[Adversarial Sample]
    For a benign input $\bm{x}$ and its predicted label $\hat{\bm{y}}$ in the classifier $f$,
    we define its adversarial counterpart as $(\bm{x}',\hat{\bm{y}}')$, where $\|\bm{x} - \bm{x}'\|_2 < \epsilon$  and  $\hat{\bm{y}} \neq \hat{\bm{y}}'$.
\end{myDef}

\begin{myDef}[Input Sensitivity (InSen)]
    With an input $\bm{x}$ and 
    its predicted label $\hat{\bm{y}}$, we define its input sensitivity as
    \begin{equation}~\label{eq:InSen}
        S(\bm{x},\hat{\bm{y}}) = \left \Vert \frac{\partial \mathcal{L}(\bm{x}, \hat{\bm{y}})}{\partial \bm{x}} \right \Vert_2,
    \end{equation}
    where $\mathcal{L}(\cdot)$ is the cross entropy loss function of $f$.
\end{myDef}

Since most of the adversarial attacks use network gradient of input images as a  guide to search perturbations~\citep{moosavi2016deepfool,carlini2017cw,rony2019ddn}, {\em the input sensitivity, which
is a form of gradient, of a benign sample and its adversarial counterpart usually have a significant difference}. The basic idea of our method is using InSen as features to design adversarial detectors. In order to prove our idea, we give following analysis.

\begin{figure}[t]
    \centering
    \includegraphics[scale=0.60]{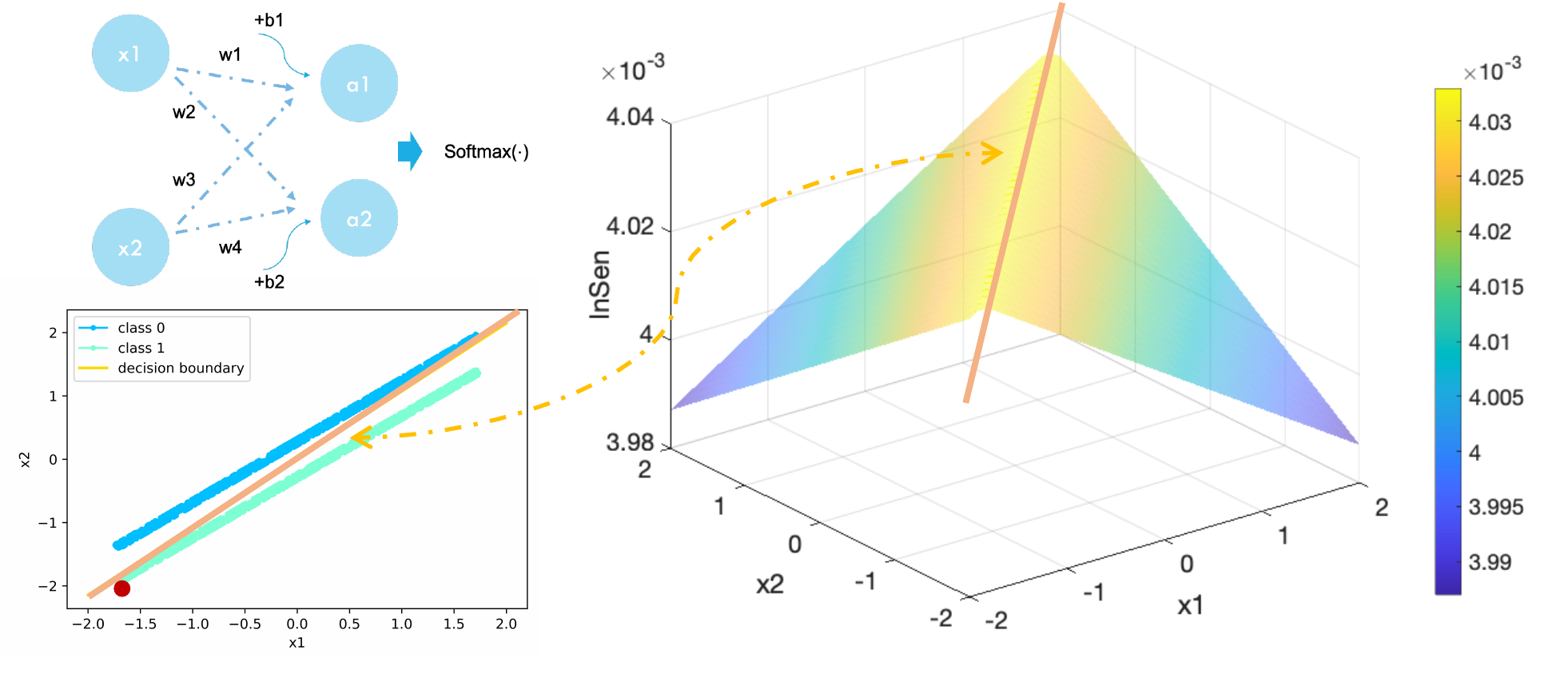}
    \caption{A simplified example of why being close to classification boundary leads to large input sensitivity. \textbf{Top left}: a toy neural network with cross entropy loss function. It takes a two dimensional input $\mathbf{X} = {(x_1, x_2)}$ with label $y \in \{0, 1\}$; \textbf{Bottom left}: The toy dataset and the extracted decision boundary (yellow line, a function of $x_1$ and $x_2$) from trained toy neural network. The data points are generated by adding random noises to uniform samples of two parallel straight lines. \textbf{Right}: The input sensitivity distribution of the toy neural network. The closer a data point is to the decision boundary (yellow line), the larger its input sensitivity (z axis) will be.}
    \label{pic:toycase}
\end{figure}

\noindent For the sake of discussion, we decompose the classifier $f$ into three parts:

1) The first part is a function that maps an input $\bm{x}$ into a representation vector $\bm{z}$, which is denoted as $\bm{z} = g(\bm{x})$.

2) The second part is a fully connected (FC) layer to calculate the prediction indexes of each classes. For the $i$-th class, the FC layer calculates an index $a_i = \bm{w}_i^\top \cdot \bm{z}$.

3) The third part is a softmax function to calculate the prediction probability of each class, which is denoted as $\bm{\hat{y}} = \sigma(\bm{a})$. For class $i$, this part is in the form of
\begin{equation}
    \hat{y}_i = \frac{\exp(a_i)}{\sum_{j=1}^N\exp(a_j)}, i \in \{1, \ldots, L\}.
\end{equation}
In other words, the classifier $f$ is in the form of
\begin{equation}
    \hat{y}_i = f_i\left(\bm{x}\right) = \sigma\left(\bm{w}_i \cdot g\left(\bm{x}\right)\right),~~i \in \{1, \ldots, L\}.
\end{equation}
Given two classes $m,n$, the label of the sample $\bm{x}$  is decided by the hyperplane,
\begin{equation}
    (\bm{w}_m - \bm{w}_n)^\top \cdot \bm{z} = 0.
\end{equation}
where $\bm{z}= g(\bm{x})$. We define a generalized distance of the sample $\bm{x}$ to the classification hyperplane as
\begin{align}\label{eq:dist_z}
    {Dist}_z\left(\bm{x}\right) = \frac{\left|(\bm{w}_m - \bm{w}_n)^\top \bm{z}\right|}{\left\|\bm{w}_m - \bm{w}_n\right\|_2}.
\end{align}
For the sample $\bm{x}$ with the predicted class $m$, following the chain rule and after rearranging some terms, its input sensitivity is in the form of
\begin{equation}\label{General Case Gradient}\small
    \begin{aligned}
        S(\bm{x}, \bm{\hat{y}}) &= \left \Vert \frac{\partial \mathcal{L}(\bm{x}, \hat{\bm{y}})}{\partial \bm{x}} \right \Vert_2
        \\&=\Bigg \Vert \sum_{p\neq m,n}{\frac{C_2}{\sum_{q}\exp{\left(a_q-a_p\right)}} } \\&+{\frac{C_1}{\exp{\underbrace{\left(\left(\bm{w}_m-\bm{w}_n\right)^\top\bm{z}\right)}_{\mathrm{Term~I}}} + \sum_{q \neq m}\exp{\left(a_q-a_n\right)}}}\Bigg \Vert_2,
    \end{aligned}
\end{equation}
where $C_1 = \left(\frac{\partial a_n}{\partial \bm{x}}-\frac{\partial a_m}{\partial \bm{x}} \right)$ and $C_2 = \left(\frac{\partial a_p}{\partial \bm{x}}-\frac{\partial a_m}{\partial \bm{x}} \right)$. Since the predicted label of $\bm{x}$ is $m$, $a_m > a_n$ and $(\bm{w}_m-\bm{w}_n)^\top\bm{z}>0$. Therefore, the {\em Term I} in Eq.~\eqref{General Case Gradient} is in direct proportion to ${Dist}_z (\bm{x})$. In other words, the InSen $S(\bm{x})$ of an example has an increasing trend when its generalized distance to the classification hyperplane decreases, \ie ${Dist}_z\left(\bm{x}\right)$ decrease. 
Therefore, we get following insights:\\
{\bf Insight I: }{The samples with shorter distances to the classification hyperplane usually have lager input sensitivities.}\\
{\bf Insight II: }{For many adversarial attack methods~\citep{moosavi2016deepfool,carlini2017cw,rony2019ddn}, the optimization object restricts size of the perturbations (such as restricting L2 metric of perturbations). Therefore, their adversarial examples are very likely lying close to the decision boundary. Thus, their input sensitivity should be larger than normal input of that class due to a small $Dist_z$.}


\begin{figure}[t]
    \centering
    \includegraphics[scale=0.26]{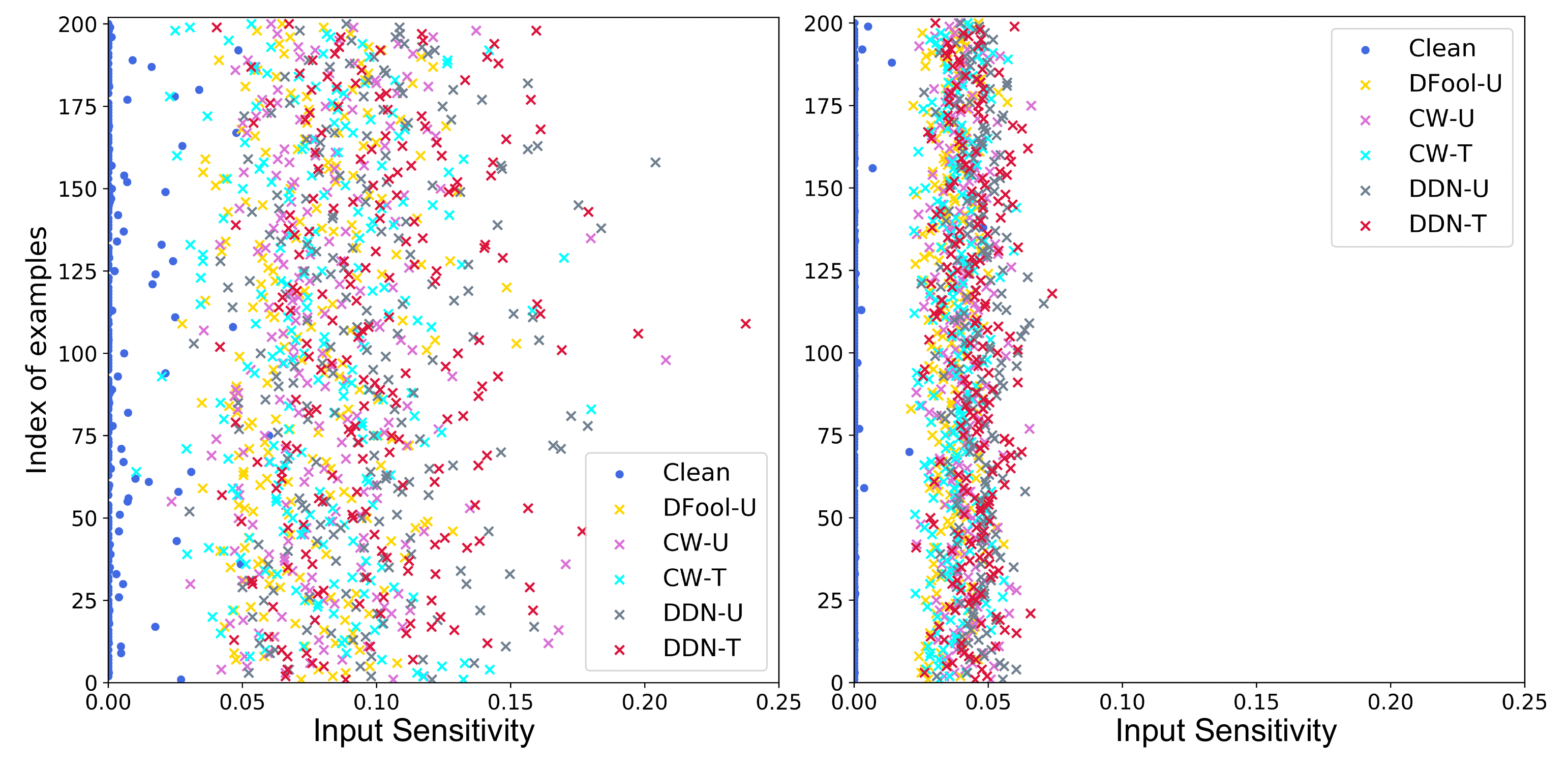}
    \caption{InSen for clean and $L_2$ adversarial examples. Left: MNIST; Right: CIFAR-10. There is a clear distinction between malicious inputs and clean inputs.}
    \label{pic:mn_grad_l2}
\end{figure}

Fig.~\ref{pic:mn_grad_l2} illustrates the $L_2$ norms of InSen for clean and malicious inputs from MNIST and CIFAR-10 (setups see the experiment section). The InSen size of clean data are very tiny, while that of the malicious data are quite gigantic. 
Fig.~\ref{pic:toycase} is an example of toy dataset on a simple single layer neural network. For example, take the red data point from class 0 with coordinate $(-1.7, -2.1)$, which is quite close to the decision boundary. From the input sensitivity distribution, we can see clearly that it is in the most yellow area indicating a large input sensitivity. These preliminary results confirmed our Insights. 


\section{Framework}
Inspired by the input sensitivity property of $L_2$ adversarial examples, we propose the {\bf G}uided {\bf C}omplementary {\bf D}efensive module (GCD). The \name model consists of two components: a detector to distinguish adversarial examples from benign examples, and a rectifier to recover adversarial examples back to its true label.

\subsection{Adversarial Example Detector}
\label{ch:detector design}
Given a set of benign examples $X = \{\bm{x}_1, \ldots, \bm{x}_N\}$, a classifier to be protected, denoted as $f: \mathbb{R}^M \rightarrow [0,1]^L$, and an adversarial attack method, we can generate a set of adversarial counterparts of $X$ as $X' = \{\bm{x}'_1, \ldots, \bm{x}'_N\}$. Combining $X$ and $X'$, we got a data set in the form of
\begin{equation}\label{}\small
  \mathcal{X} = \Big\{\left(\bm{x}_1, \bm{l}_1\right), \ldots, \left(\bm{x}_N, \bm{l}_N\right), \left(\bm{x}'_1, \bm{l}'_1\right), \ldots, \left(\bm{x}'_N, \bm{l}'_N\right)\Big\},
\end{equation}
where $\bm{l} = (l^+, l^-)$ is a one-hot coding of benign/adversarial label. $\bm{l}_n = (1,0)$ for benign examples and $\bm{l}'_n = (0,1)$ for adversarial examples.

The \name model uses the data set $\mathcal{X}$ to train two sub-detectors.\footnote{See supplementary for full neural network structure of detectors.} 
Firstly, we can directly feed gradients of $\bm{x}$ from base model being protected to the detector. As shown in Fig.\ref{pic:mn_grad_l2}, the only thing the detector need to learn is to calculate $L_2$ norm of the input and find a suitable divider between large input sensitivity and normal input sensitivity, \ie
\begin{equation}\label{eq:classifier_2}
  (p^+, p^-) = DET_{IS}\big(G(\bm{x}, \hat{\bm{y}})\big),
\end{equation}
where $G(\bm{x}, \hat{\bm{y}})$ is the gradient map of sample $\bm{x}$ and $p^+ \in [0,1]$ is the probability of the input being benign, $p^- \in [0,1]$ is the probability of being adversarial ones, and $p^+ + p^- = 1$. The image label $\hat{\bm{y}}$ is generated by the protected base model.

\begin{figure}[t]
    \centering
    \includegraphics[scale=0.26]{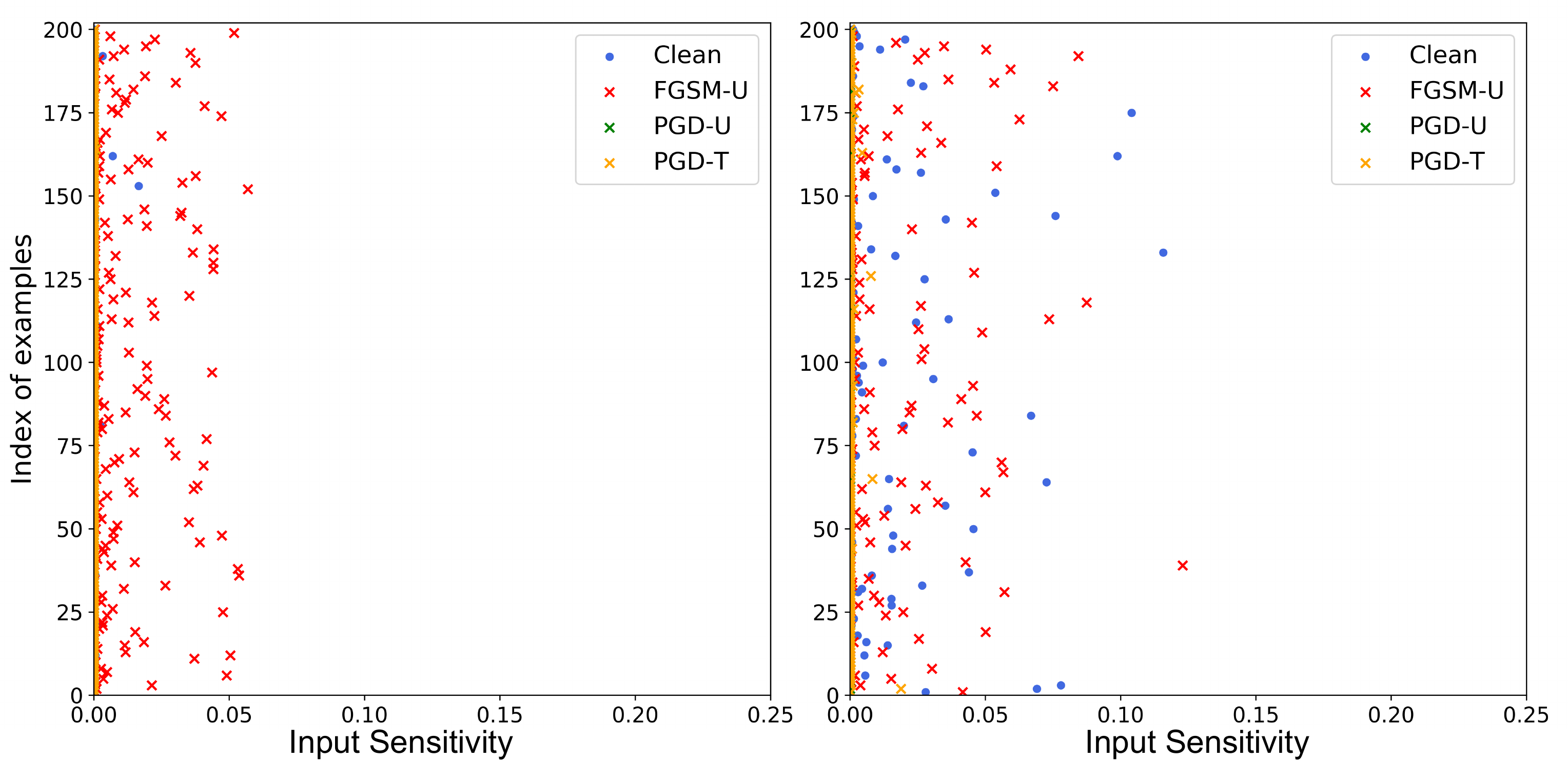}
    \caption{InSen for clean and $L_\infty$ adversarial examples. Left: MNIST; Right: CIFAR-10. There isn't any clear separation between malicious inputs and clean inputs.}
    \label{pic:lfinsen}
\end{figure}

However, we notice that some adversarial examples generated by $L_\infty$ attacks may evade from our detection, this may be caused by the fact that their $L_2$ norm are not directly bounded and optimized. They are usually far from the classification boundary~\citep{Yu_2019_NeurIPSweakness} and have similar InSen as normal inputs~\ref{pic:lfinsen}. But one weakness of not constraining $L_2$ norm is that the adversarial examples will be distinguishable in pixel domain. Thus, we use a second sub-detector adopts $\bm{x}_n$ (or $\bm{x}'_n$) as inputs of a convolutional neural network (CNN) to predict their benign/adversarial labels, \ie
\begin{equation}\label{eq:classifier_1}
  (p^+, p^-) = DET_{org}\big(\bm{x}\big),
\end{equation}


Though there are preliminary work~\citep{DBLP:journals/corr/GongWK17, 2018qinghua_imagegrad} that has used similar detectors setup as our sub-detectors, none of them have discovered the complementary power of these two sub-detectors empirically and theoretically. That is to say, $DET_{IS}$ is suitable for detecting $L_2$ attacks while $DET_{org}$ performs well on $L_\infty$ attacks. We further propose a information entropy based ensemble mechanism to gain this complementary advantage.

Formally, given an input, we denote the outputs of $DET_{org}$ and $DET_{IS}$ as $\bm{p}_{o} = (p^+_o, p^-_{o})$ and $\bm{p}_{s} = (p^+_{s}, p^-_{s})$, respectively. For the two predicted labels, \name estimates the information entropy they could offer as
\begin{equation}\label{}
\begin{aligned}
  {H}_{o} &= - \left[p^+_o \log_2(p^+_o) + p^-_o \log_2(p^-_o)\right], \\
  {H}_{s} &= - \left[p^+_s \log_2(p^+_s) + p^-_s \log_2(p^-_s)\right].
\end{aligned}
\end{equation}
The prediction from the classifier that is more confident is adopted as the final prediction of \name, \ie
\begin{equation}\label{eq:H_combine}
  (p^+, p^-) = \left\{\begin{aligned}
  \bm{p}_o &~~~ \mathrm{if}~~ {H}_{o} < {H}_{s} \\
  \bm{p}_s &~~~ \mathrm{if}~~ {H}_{o} \geq {H}_{s}
\end{aligned}\right. .
\end{equation}

The \name model considers the images with $p^+ \geq p^-$ as benign examples, of which the labels of the original image classification task are directly assigned by the classifier $f$. On the contrary, the images with $p^+ < p^-$ are considered as adversarial examples. Their predicted label should be recovered by a rectifier.

\subsection{Adversarial Example Rectifier}

\begin{table}[t]
  \begin{center}
    \begin{tabular}{|c|ccc|}
    \hline
    \multicolumn{1}{|c|}{\multirow{2}[0]{*}{Removed Pixels Percent}} & \multicolumn{3}{c|}{Attack Success Rate} \\
\cline{2-4}      & Deepfool & CW & DDN \\
    \hline
    top 0\% & 1.000  & 1.000  & 1.000  \\
    top 5\% & 0.637  & 0.665  & 0.656  \\
    top 10\% & 0.780  & 0.794  & 0.785  \\
    \hline
    \end{tabular}
  \end{center}
    \caption{Attack success rates of adversarial examples excluded top 5\% and 10\% high saliency pixels according to detector gradients.}%
    \label{tab:toy_exp} \centering \small
\end{table}%

The design of the rectifier is inspired by the hypothesis that the gradient map of detectors can be viewed as a form of saliency which is a clear evidence that, the gradient map of detectors can indicate the image area being polluted. Therefore, we conjecture if we remove the polluted pixels guided by the corresponding detector's saliency map, the attack might be alleviated. 

Table~\ref{tab:toy_exp} are results of an simple experiment for verification. We use three methods, Deepfool~\citep{moosavi2016deepfool}, CW~\citep{carlini2017cw} and DDN~\citep{rony2019ddn}, to generate adversarial examples for a VGG11 network, and set the pixels with top 5\% and 10\% biggest $\left|\partial \mathcal{L}_{DET}/\partial x_i\right|$ as zero respectively. As shown in the table, when using original adversarial examples to attack the VGG11 network, the success rate is 100\%, but when we use adversarial examples without top 5\% high saliency pixels to attack, the success rate falls to only 63.7\%. This experiment results align well with our conjecture that gradient map from detectors can be used as indexes to pick polluted pixels out of an adversarial input. When the ratio of removed pixels increases, the success rate increases again as essential information needed for classification misses too much. This requires us to design an elaborate approach to pick out the polluted pixels. 

We define an adversarial example that is detected successfully as a matrix $\bm{X}$, where the element $x_{a,b}$ is the pixel at the location $(a,b)$\footnote{For the sake of discussion, we use the signal channel image as an example to introduce our rectifier.}. If the detection result comes from $\bm{p}_o$, gradients from $DET_{org}$ will be used. Otherwise, if it is from $\bm{p}_s$, gradients from $DET_{IS}$ will be used.
We select a pixel as a suspect when its detector gradient is larger than a threshold,
\begin{equation}\label{eq:masking_rate}
    \left|\frac{\partial \mathcal{L}}{\partial x_{a,b}}\right| > \alpha \cdot \Big(G_{max}- G_{min}\Big) + G_{min},
\end{equation}
where $G_{max}$ and $G_{min}$ are the maximum and minimum values of pixels' sensitivity in the image $\bm{X}$, and $\alpha$ is a percent parameter to control the ratio of suspect pixels.

For a suspect pixel, it will be suppressed randomly according to two random numbers $x_{ran}$ and $u_{ran}$,
\begin{equation}\label{}\small
    x_{a,b} : = \left\{\begin{aligned}
  & x_{a,b} &~~~ \mathrm{if}~~ u_{ran} = 0 \\
  & x_{ran} &~~~ \mathrm{if}~~ u_{ran} = 1
\end{aligned}\right.,
\end{equation}
where $u_{ran}\in \{0,1\}$ follows a Bernoulli distribution. If $u_{ran} = 0$, we keep $x_{a,b}$ as its raw value, otherwise we set it as a random value $x_{ran} \in [0,255], x_{ran} \in \mathbb{N}$. $x_{ran}$ follows a normal distribution ${N}(\sigma, \mu)$, where $\sigma, \mu$ are set as the mean and standard deviation of pixels in $\bm{X}$.

In this way, we can generate various randomized duplicates for each adversarial example $\bm{X}_i$ as a dataset $\mathcal{X}_{ran}$. Then we fine tune the original image classifier as the rectifier $f_{rec}$. 
Compared with the original image classifier $f$, $f_{rec}$ is more robust to adversarial examples. However, its performance on benign examples is slightly worse. To combine the advantages of both $f$ and $f_{rec}$, we only use $f_{rec}$ to classify the adversarial examples picked out by the detector, and handle clean inputs with $f$.

\section{Experiment}

\subsection{Experiment Setup}
In the experiments, we test our $\name$ module on MNIST\citep{lecun1998mnist}, CIFAR-10\citep{krizhevsky2009cifar} and a subset of ImageNet\citep{imagenet_cvpr09}. We select images from 20 object classes with each class containing 1,040 training samples and 260 testing samples. All the pixels' scale is in [0,1].
We use VGG11~\citep{vgg11} as base classifier on the MNIST and CIFAR-10 datasets. Our base classifiers obtain test accuracy of 99.17\% and 84.41\% on MNIST and CIFAR-10 respectively. For ImageNet subset, we use a pre-trained ResNet152~\citep{resnet} with 95.25\% prediction accuracy on the test set.

Experiments are conducted on five extensively used and state of the art attacks, which are FGSM~\citep{fgsm},  Deepfool (DFool) ~\citep{moosavi2016deepfool}, PGD~\citep{madry2017pgd}, CW~\citep{carlini2017cw} and DDN~\citep{rony2019ddn}. For each algorithm, we evaluate our model on two types of threats, untargeted and targeted~\citep{akhtar2018threat}.

1) {\em Untargeted Attack:} For a classifier $f: \mathbb{R}^M \rightarrow [0,1]^L$ equipped with a detector, an input image can be classified into $L+1$ classes, \ie $L$ object classes and one adversarial class $y_{adv}$. For an image with an original prediction of $\hat{y}_{true}$, its untargeted adversarial counterpart is successful only when the perturbed predicted label $\hat{y}'$ satisfies $\hat{y}' \neq \hat{y}_{true}$ and $\hat{y} \neq y_{adv}$.

2) {\em Targeted Attack:} For an image with an original predicted label $\hat{y}_{true}$, its targeted adversarial counterpart is successful only when the perturbed label $\hat{y}'$ satisfies $\hat{y}' = y_{target}, y_{target} \neq \hat{y}_{true}$. 

We mark the untargeted version of attack algorithms with the postfix ``-U'', and  the targeted version with ``-T''.\footnote{FGSM and Deepfool only have untargeted version, while PGD, CW, DDN have both targeted and untargeted versions.}

We mainly evaluate our module under two attack scenarios, white-box and grey-box to verify our analysis that attacks guided by gradients information from the base model are intrinsically different to be detected and recovered.

1) {\em Grey-box Attack}: The attacker has knowledge of the network parameters and structures of the target model but is not aware of the existence of defensive mechanisms.

2) {\em White-box Attack}: The attacker is aware that the target model is under the protection, and has full access to the target model and the defensive module.

\subsection{Performance of Detector}

In this section we evaluate our detector in grey and white-box scenarios under the metric of AUC score. All the detectors use a VGG11 unless specified. Here, we compare the proposed detector $DET_{\name}$ with six baselines:

1) $DET_{org}$ uses original image as input to train a detector~\citep{DBLP:journals/corr/GongWK17} (see Eq.~\eqref{eq:classifier_1}).

2) $DET_{sal}$ uses saliency maps of network interior layers as features to train a detector, which is first proposed by Zhang \etal~\citep{2018qinghua_imagegrad}.

3) $DET_{IS}$ only uses input sensitivity to train a detector (see Eq.~\eqref{eq:classifier_2}).

4) $DET_{TWS}$~\citep{tws} utilizes two seemingly paradoxical criteria of
benign images to detect adversarial examples, with the assumption
that an adversarial example cannot satisfy both two criteria. 

5) $DET_{PD}$~\citep{pixeldefend} utilizes a PixelCNN network to extract
features of examples, and adopts a test statistic to decide whether
an input is an adversarial example. 

6) $DET_{MDS}$~\citep{mds} utilizes Mahalanobis Distance-based Score to measure different distributions of adversarial and benign examples in a network’s interior-layers' outputs. 


\subsubsection{The Grey-box Attacks}

\begin{table*}[t] \footnotesize
  \begin{center}

    \begin{tabular}{|c|c|c|c|c|c|c|c|c|c|}
    \hline
    Dataset & Attacker & F     & $DET_{org}$   & $DET_{sal}$   & $DET_{IS}$    & $DET_{PD}$    & $DET_{TWS}$   & $DET_{MDS}$   & Our \\
    \hline
    \hline
    \multirow{8}[2]{*}{\begin{sideways}MNIST\end{sideways}} & FGSM-U & 0.65  & \textbf{1.00} & \textbf{1.00} & 0.99  & 1.00  & 0.73  & 0.81  & \textbf{1.00} \\
          & PGD-U & 0.00  & \textbf{1.00} & \textbf{1.00} & 0.89  & 1.00  & 0.75  & 0.89  & \textbf{1.00} \\
          & PGD-T & 0.00  & \textbf{1.00} & \textbf{1.00} & 0.72  & 1.00  & 0.93  & 0.90  & \textbf{1.00} \\
          & DFool-U & 0.00  & \textbf{1.00} & \textbf{1.00} & \textbf{1.00} & 0.98  & 0.99  & 0.87  & \textbf{1.00} \\
          & CW-U  & 0.00  & 0.99  & \textbf{1.00} & \textbf{1.00} & 0.93  & 0.95  & 0.89  & \textbf{1.00} \\
          & CW-T  & 0.00  & \textbf{1.00} & \textbf{1.00} & \textbf{1.00} & 0.98  & 0.99  & 0.93  & \textbf{1.00} \\
          & DDN-U & 0.00  & \textbf{1.00} & \textbf{1.00} & \textbf{1.00} & 0.98  & 0.80  & 0.89  & \textbf{1.00} \\
          & DDN-T & 0.00  & \textbf{1.00} & \textbf{1.00} & \textbf{1.00} & \textbf{1.00} & \textbf{1.00} & 0.95  & \textbf{1.00} \\
    \hline
    \hline
    \multirow{8}[2]{*}{\begin{sideways}CIFAR-10\end{sideways}} & FGSM-U & 0.19  & \textbf{1.00} & \textbf{1.00} & 0.95  & 0.99  & 0.52  & 0.83  & \textbf{1.00} \\
          & PGD-U & 0.10  & \textbf{1.00} & \textbf{1.00} & 0.86  & 0.99  & 0.52  & 0.76  & \textbf{1.00} \\
          & PGD-T & 0.18  & \textbf{1.00} & \textbf{1.00} & 0.64  & 0.99  & 0.58  & 0.71  & \textbf{1.00} \\
          & DFool-U & 0.00  & 0.50  & 0.50  & \textbf{0.98} & 0.77  & 0.83  & 0.93  & \textbf{0.98} \\
          & CW-U  & 0.00  & 0.50  & 0.50  & \textbf{0.98} & 0.78  & 0.90  & 0.93  & \textbf{0.98} \\
          & CW-T  & 0.00  & 0.51  & 0.51  & \textbf{0.98} & 0.84  & 0.94  & 0.94  & \textbf{0.98} \\
          & DDN-U & 0.00  & 0.50  & 0.50  & \textbf{0.99} & 0.70  & 0.91  & 0.93  & \textbf{0.99} \\
          & DDN-T & 0.00  & 0.51  & 0.53  & \textbf{0.99} & 0.81  & 0.96  & 0.94  & \textbf{0.99} \\
    \hline
    \hline
    \multirow{8}[2]{*}{\begin{sideways}ImageNet\end{sideways}} & FGSM-U & 0.27  & \textbf{1.00} & \textbf{1.00} & 0.97  & 0.99  & 0.54  & 0.80  & 0.99 \\
          & PGD-U & 0.00  & \textbf{1.00} & \textbf{1.00} & 0.88  & 0.98  & 0.54  & 0.79  & \textbf{1.00} \\
          & PGD-T & 0.01  & \textbf{1.00} & \textbf{1.00} & 0.71  & 0.98  & 0.60  & 0.75  & \textbf{1.00} \\
          & DFool-U & 0.01  & 0.50  & 0.50  & \textbf{0.99} & 0.80  & 0.79  & 0.91  & \textbf{0.99} \\
          & CW-U  & 0.00  & 0.50  & 0.50  & \textbf{0.98} & 0.82  & 0.86  & 0.90  & \textbf{0.98} \\
          & CW-T  & 0.00  & 0.50  & 0.50  & \textbf{0.98} & 0.79  & 0.90  & 0.91  & \textbf{0.98} \\
          & DDN-U & 0.00  & 0.50  & 0.50  & \textbf{1.00} & 0.71  & 0.91  & 0.93  & \textbf{1.00} \\
          & DDN-T & 0.00  & 0.50  & 0.50  & \textbf{1.00} & 0.84  & 0.91  & 0.94  & \textbf{1.00} \\
    \hline
    \end{tabular}%
    \end{center}
  \caption{Summary of the detectors performance (AUC score) in grey-box scenario. F stands for base model accuracy under attacks.}
  \label{tab:grey-box evaluation}%

\end{table*}%

For grey-box attacks, the attacker has full access to base classifiers' structures and parameters, but the attacker does not know the existence of our detector. Adversarial examples are not optimized for the detector. This is a favorable condition for our detector, since the enemies are exposed and we are in a shelter. 
Tab.~\ref{tab:grey-box evaluation} shows AUC score of grey-box attacks. 
As shown in the table, 
we observe that:

1) The $DET_{\name}$ achieves above 98\% AUC for all datasets and all attacks.


2) On harder datasets CIFAR-10 and ImageNet, $DET_{IS}$ only shows good performance under $L_2$ attacks, which verifies Insight I we proposed in Sec.~\ref{sec:motivation}.

3) For FGSM and PGD attacks, the performance of $DET_{org}$ is better than $DET_{IS}$. As we claimed in \ref{ch:detector design}, the reason lies behind is they are not bounded by $L_2$ norm, while they are easily detected by $DET_{org}$ as the pixel space distortion is big enough. 

4) $DET_{\name}$ benefits from both $DET_{IS}$ and $DET_{org}$ using Eq.~\eqref{eq:H_combine} and achieves the best average performance (estimated by Win Count), which is close to the better one between $DET_{IS}$ and $DET_{org}$ for most of the time.

5) $DET_{\name}$ outperforms all three recent state-of-the-art detectors by a large margin.

\subsubsection{The White-box Attacks}


    

\begin{table*}[htbp] \footnotesize
  \begin{center}
    \begin{tabular}{|c|c|c|c|c|c|c|c|c|c|}
    \hline
    \multirow{2}[4]{*}{Attacker} & \multicolumn{3}{c|}{MNIST} & \multicolumn{3}{c|}{CIFAR10} & \multicolumn{3}{c|}{ImageNet} \\
\cline{2-10}    & NoDET & $DET_{IS}$ & Our & NoDET & $DET_{IS}$ & Our & NoDET & $DET_{IS}$ & Our\\
    \hline
    \hline
    FGSM-U & 0.271 & 0.012 & \textbf{0.000} &0.315 & 0.009 & \textbf{0.000} &0.608 & 0.566 & \textbf{0.180} \\

    PGD-U & 1.000 & 0.871 & \textbf{0.263} & 0.945 & 0.805 & \textbf{0.143} & 1.000 & \textbf{0.359} & 0.425  \\
    PGD-T & 1.000 & 1.000 & \textbf{0.437} & 0.957 & 0.734 & \textbf{0.463} & 1.000 & \textbf{0.456} & 0.485  \\
    
    DFool-U & 1.000 & 0.828 & \textbf{0.286}  & 1.000 & 0.738 & \textbf{0.394}  & 0.995 & \textbf{0.087} & \textbf{0.087}  \\
    CW-U & 1.000 & \textbf{0.000} & 0.008 & 1.000 & \textbf{0.009} & 0.023 & 1.000 & \textbf{0.009} & 0.023  \\
    CW-T & 1.000 & \textbf{0.180} & 0.471 & 1.000 & 0.556 & \textbf{0.321}  & 0.995 & 0.056 & \textbf{0.015}  \\
    DDN-U & 1.000 & \textbf{0.004} & {0.941}  & 1.000 & 0.019 & \textbf{0.014}  & 1.000 & 0.159 & \textbf{0.092}  \\
    DDN-T & 1.000 & 0.808 & \textbf{0.392}  & 1.000 & 1.000 & {1.000}  & 1.000 & 0.338 & \textbf{0.326}  \\
    \hline
    \end{tabular}%
  \end{center}
  \caption{Attack success rate under white-box attacks. Smaller is better.}
  \label{tab:white-box_t}%
\end{table*}%


In this scenario, attackers have full knowledge of our defensive model and they can generate adversarial examples accordingly. We evaluate our model under targeted and untargeted attacks. For targeted attacks, we use the method proposed by Carlini\&Wagner~\citep{carlini2017adversarial}, which treats our \name model as a $L+1$ classifier, \ie $L$ natural classes and an adversarial class. For the untargeted attacks, we optimize a combined cross entropy loss function to evade our proposed defense,
\begin{equation}\label{eq:combined object}
\begin{aligned}
  \mathcal{L}_{com} &= \mathcal{L}_{adv} + \beta\mathcal{L}_{\name}
\end{aligned}
\end{equation}
where $\mathcal{L}_{com}$ is the combined cross entropy loss, $\mathcal{L}_{adv}$ is the loss for image label, $\mathcal{L}_{\name}$ is the loss for our detector and $\beta$ is a constant to balance the weight of detector loss in the combined object. See details in supplementary. 

As suggested by Carlini and Wagner~\citep{carlini2017adversarial}, a successful white-box attack should not be detected by the detector as well as fooling the base classifier. The adversarial examples that cannot bypass the detector or fail to fool the classifier are not successful.

Therefore, the AUC score of binary classification is no longer suitable for evaluating the worst case performance of our detector. Instead, we use success rate of adversarial examples to evaluate performance of our detectors. We consider an attacker fails when its attack success rates is less than 50\%. From Tab.~\ref{tab:white-box_t}, we can see that,

1) Even in the worst case, our model \name can still successfully defeat FGSM, PGD, and CW. 

2) Our model achieves full success on ImageNet subset. The defense fails when it is tested against untargeted DDN on MNIST and targeted DDN on CIFAR-10. We highlighted these failed defenses in blue for further analysis.


\begin{table}[t]\footnotesize
  \begin{center}
    \begin{tabular}{|c|c|c|c|}
    \hline
    Attacker  & DDN-U  & DDN-T \\
    \hline
    Without detector & 0.787 &  0.45 \\
    \hline
    With Our $DET_{\name}$ & 10.252 & 0.618 \\
    \hline
    \end{tabular}%
  \end{center}
  \caption{The L2 perturbations of the three {``failed''} attacks. It is clear that the perturbation required for attacking protected models is larger than that for vanilla models.}
  \label{tab:perturbations_l2}%
\end{table}%

\begin{figure}
  \begin{center}
    \subfigure[Untargeted DDN for MNIST]{\includegraphics[width=0.5\textwidth]{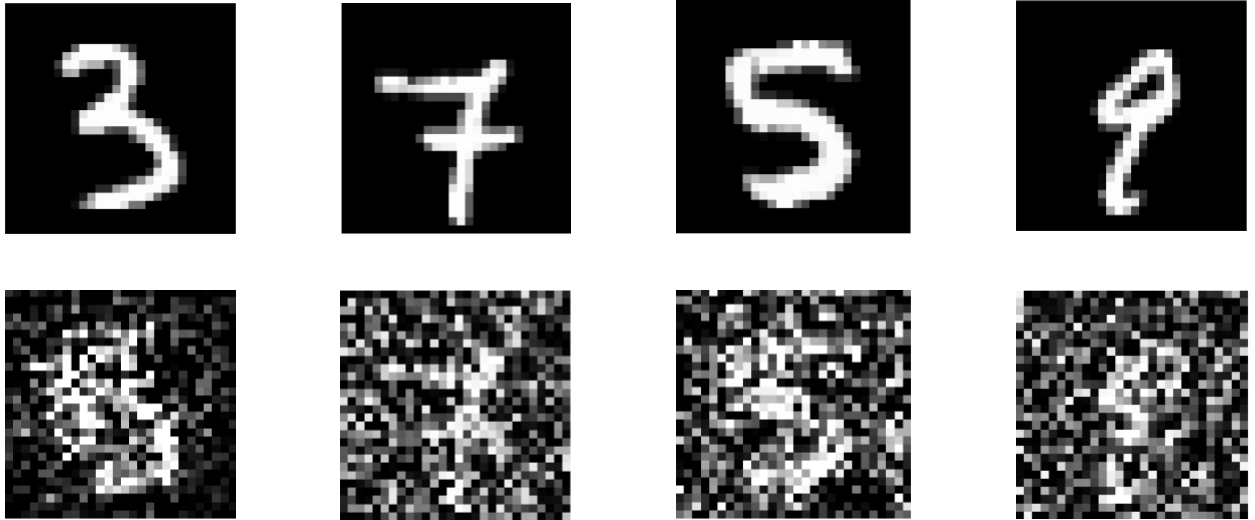}} \\
    \subfigure[Targeted DDN for CIFAR-10]{\includegraphics[width=0.5\textwidth]{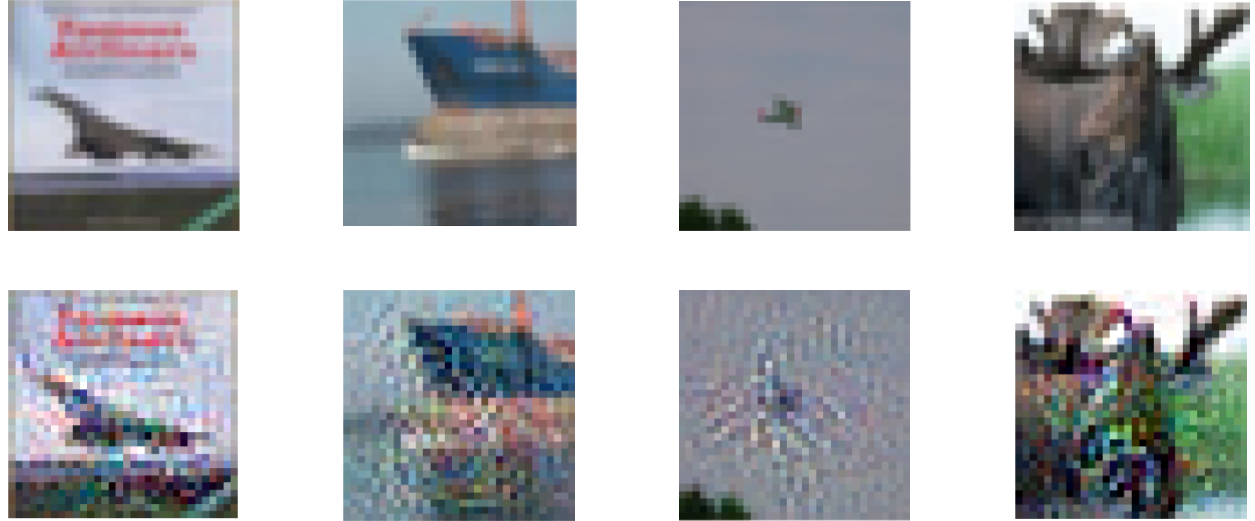}} \\
  \end{center}
  \caption{Example perturbed images from ``failed'' cases. The resulted outputs are highly noisy, which is another indicator that our method does increase the attacking cost even being bypassed. First row in each sub-figure is clean and the second row is perturbed.} \label{fig:noise_examples}
\end{figure}

However, does our defense really fail? We need to open the box to look into some details. Tab.~\ref{tab:perturbations_l2} shows the $L_2$ perturbations of these three {``failed''} cases. Obviously, these attacks evade our defense to some extent at the cost of significantly increased $L_2$ perturbations. Fig.~\ref{fig:noise_examples} also gives example perturbed images from these three attacks. The noise can be observed easily. Therefore, we cannot say these samples are truly ``successful''.

Besides, for targeted DDN, of which the L2 perturbations is pretty small, we still cannot say the attack is ``success'' due to huge computing resource consumption. In our experiments, it spends about 40 hours to train the targeted DDN attacker over 10,000 CIFAR10 dataset. However, it only takes one hour to train DDN on a vanilla classifier without protection. We believe the time consumption will become unacceptable for training DDN attacker that can fool $\name$ when it comes to a huge real-world image dataset.

\begin{table*}[t] \tiny
  \begin{center}
  
    \begin{tabular}{|c|c|c|c|c|c|c|c|c|c|c|c|c|c|c|c|c|c|c|}
    \hline
          & \multicolumn{6}{c|}{MNIST}                    & \multicolumn{6}{c|}{CIFAR-10}                 & \multicolumn{6}{c|}{ImageNet} \\
    \hline
          & Our   & PD    & DDN   & PGD   & TVM   & F     & Our   & PD    & DDN   & PGD   & TVM   & F   & Our   & PD    & DDN   & PGD   & TVM   & F \\
    \hline
    \hline
    Clean & 0.91  & 0.90  & 0.86  & 0.84  & 0.67  & \textbf{0.92} & 0.82  & 0.79  & 0.75  & 0.64  & 0.35  & \textbf{0.86} & 0.86  & 0.66  & 0.78  & 0.72  & 0.75  & \textbf{0.95} \\
    FGSM-U  & \textbf{0.85} & 0.75  & 0.82  & 0.82  & 0.49  & 0.56  & \textbf{0.64} & 0.36  & 0.55  & 0.43  & 0.29  & 0.24  & \textbf{0.67} & 0.47  & 0.49  & 0.47  & 0.36  & 0.44 \\
    PGD-U & \textbf{0.84} & 0.64  & 0.80  & 0.81  & 0.57  & 0.27  & \textbf{0.77} & 0.30  & 0.37  & 0.35  & 0.32  & 0.08  & \textbf{0.77} & 0.70  & 0.38  & 0.47  & 0.66  & 0.02 \\
    PGD-T & \textbf{0.89} & 0.86  & 0.84  & 0.87  & 0.53  & 0.66  & \textbf{0.74} & 0.62  & 0.33  & 0.48  & 0.32  & 0.05  & \textbf{0.77} & 0.66  & 0.29  & 0.51  & 0.73  & 0.00 \\
    DFool-U & \textbf{0.90} & 0.88  & 0.26  & 0.76  & 0.65  & 0.00  & \textbf{0.74} & 0.71  & 0.19  & 0.29  & 0.34  & 0.00  & \textbf{0.82} & 0.58  & 0.37  & 0.35  & 0.75  & 0.01 \\
    CW-U  & \textbf{0.90} & 0.88  & 0.70  & 0.73  & 0.66  & 0.00  & \textbf{0.75} & 0.73  & 0.70  & 0.63  & 0.34  & 0.00  & \textbf{0.80} & 0.64  & 0.50  & 0.53  & 0.74  & 0.00 \\
    CW-T  & \textbf{0.89} & 0.86  & 0.72  & 0.53  & 0.65  & 0.00  & 0.74  & \textbf{0.75} & 0.45  & 0.46  & 0.33  & 0.00  & 0.73  & 0.61  & 0.40  & 0.39  & \textbf{0.75} & 0.00 \\
    DDN-U & \textbf{0.90} & 0.89  & 0.74  & 0.76  & 0.66  & 0.00  & \textbf{0.75} & 0.74  & 0.66  & 0.52  & 0.34  & 0.00  & \textbf{0.83} & 0.60  & 0.56  & 0.44  & 0.75  & 0.03 \\
    DDN-T & \textbf{0.90} & 0.89  & 0.59  & 0.64  & 0.65  & 0.00  & \textbf{0.75} & \textbf{0.75} & 0.53  & 0.43  & 0.34  & 0.00  & \textbf{0.80} & 0.60  & 0.50  & 0.39  & 0.74  & 0.00 \\
    \hline
    \end{tabular}%
    \end{center}
  \caption{Classification Accuracy of our \name module. F stands for the base model being protected. PGD and DDN are two adversarial trained baselines. TVM stands for Total Variance Minimization and PD stands for PixelDefend, which are two famous rectification methods.}
\label{tab:end2end_performance}%

\end{table*}%

\subsection{Performance of Rectifier}
 
Next, we test the performance of the rectifier. The implementation of the rectifier includes two steps: 1) generating images with detector gradients guided random mask; 2) using the masked images to fine tune classifier. These two steps both contribute to the rectifier's performance improvement. Therefore, we conduct analysis on these two steps respectively.


\begin{figure}
  \centering
{\includegraphics[width=0.65\textwidth]{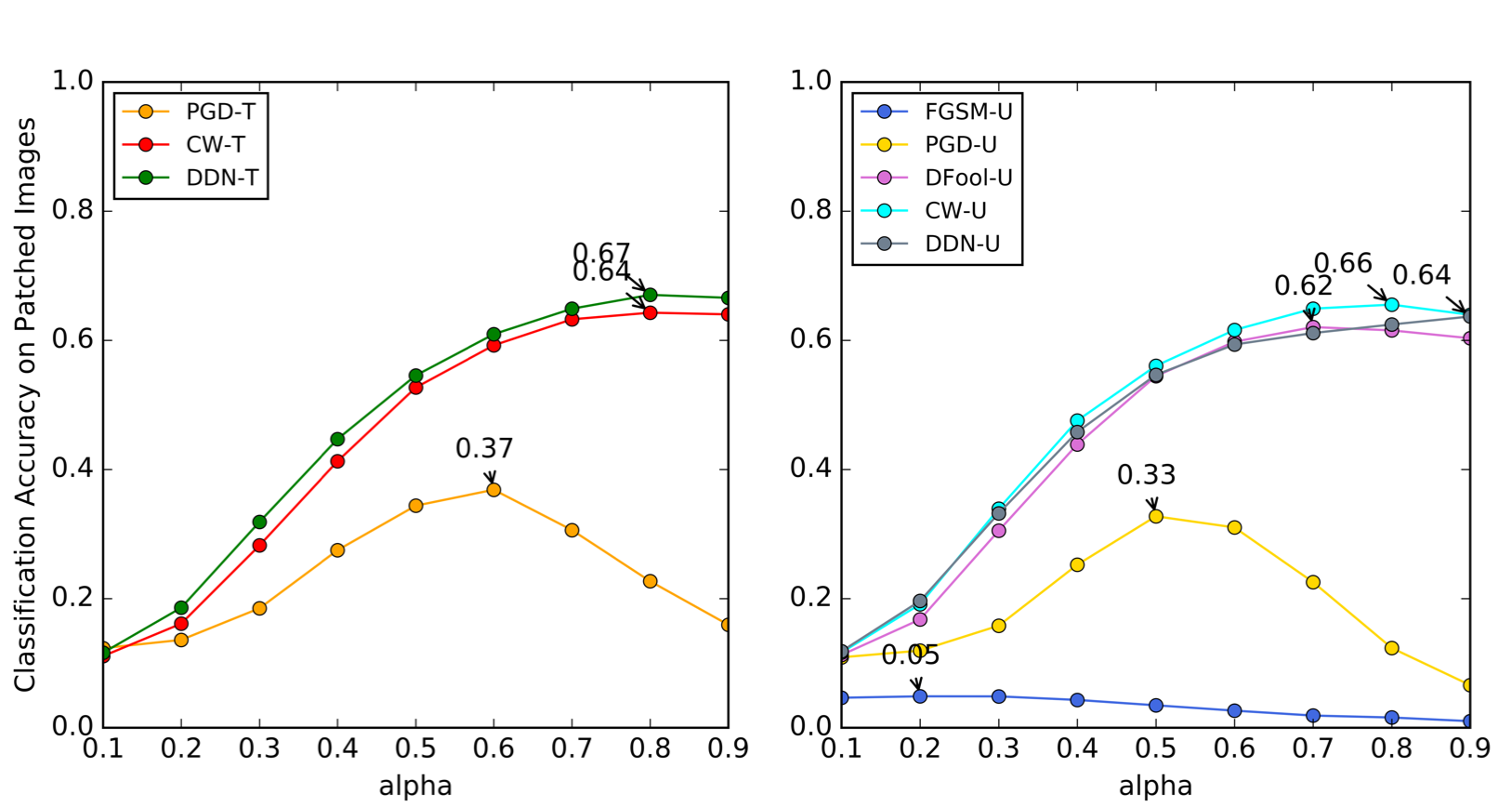}} 
  \caption{Influence of different masking rate on the rectifier's performance. Left: Targeted. Right: Untargeted.}
    \label{fig:alpha curve}
\end{figure}

We first evaluate how different masking rate influences the performance of InSen guided rectification. Fig.~\ref{fig:alpha curve} shows the rectifier performance on adversarial examples of our ImageNet subset. We vary the parameter $\alpha$, which controls the masking rate, in Eq.~\eqref{eq:masking_rate} from 10\% to 90\%. Obviously, neither too large nor too small $\alpha$ is a good choice. Too small $\alpha$ cannot eliminate the influence of adversarial perturbations; too large $\alpha$ may cause loss of too much image information. In untargeted scene, we use 0.55 for MNIST, 0.65 for CIFAR-10, 0.45 for ImageNet subset. In targeted scene, we use 0.6 for MNIST, 0.7 for CIFAR-10, 0.35 for ImageNet subset.

Secondly, we mask adversarial examples and uses the masked examples to fine tune the base classifier as a rectifier. Tab.~\ref{tab:end2end_performance} lists the adversarial example classification performance of the rectifiers for different datasets. As shown by the table, our recovery accuracy for the adversarial examples surpass all the baselines including those adversarial trained models (PGD, DDN)\citep{madry2017pgd, rony2019ddn} and some famous rectification method (PD, TVM)\citep{pixeldefend, tvm}.

It is worthy of mentioning that,

1) Here we train our PGD adversarial classifier using the same parameters reported in~\citep{madry2017pgd} on MNIST, CIFAR-10. Their adversarial trained Wide ResNet/simple ResNet can only achieve around 45\% accuracy on CIFAR-10, while our VGG11 classifier could achieve more than 74\%. The situation is same for MNIST. This also dampens their claim on the influence of model capacity to the robustness.  

2) Our VGG11 achieves 99\% and 88\% on MNIST and CIFAR-10 respectively, while adversarial trained VGG11 using DDN~\citep{rony2019ddn} only achieves 87\% and 67\%.

3) On ImageNet subset, our rectifier using wide ResNet shows significant superiority over other baselines. 

\section{Related Work}

Previous research on finding adversarial examples assume attackers have full access to the models called ``white box" attacks. Many widely applied attacks leverage gradient of the model to guide their search towards efficient perturbations~\citep{carlini2017cw,bim,fgsm,su2019onepixel,moosavi2016deepfool,rony2019ddn,madry2017pgd}. In a more realistic setting, direct access to the target model is not always available to attackers. Thus, subsequent work also focuses on ``black box" attacks~\citep{papernot2016transferability,papernot2017practical,alzantot2019genattack,pmlr-v97-guo19a}. From an optimization perspective, to search for a suitable adversarial perturbation is to solve a certain object function with constrained noise payload under some measures like $L_0, L_2, L_\infty$.

The methods proposed to mitigate the threats of adversaries could mainly be divided into two categories. One is to make robust predictions against adversaries directly. This could be done in various approaches including robust optimization, model architecture modification and inference time reconstruction of inputs.
1) Robust optimization~\citep{madry2017pgd,fgsm} mainly augments the training data by adding adversarial examples to train a robust classifier. It usually requires online update of training data consuming exorbitant time and computational resources which makes it difficult at ImageNet scale~\citep{scale17}.
2) Modifying model architectures~\citep{BuckmanRRG18, DhillonALBKKA18, SamangoueiKC18, DBLP:conf/iclr/XieWZRY18, DBLP:conf/iclr/SongKNEK18} typically means manipulating the model gradients to make it intractable at inference time for adversarial attacks or use sub networks like defensive distillation~\citep{DBLP:conf/sp/PapernotM0JS16}. These methods have also been proved to be ineffective~\citep{DBLP:conf/icml/AthalyeC018,DBLP:journals/corr/CarliniW16}. 3) Reconstruction of inputs~\citep{GuoRCM18} during inference time could remove input perturbations to some extent so that the input could be classified correctly. This method could also be bypassed with BPDA~\citep{DBLP:conf/icml/AthalyeC018} which approximates the gradient of the target network.

Considering the hardness of counteracting attacks directly, another main stream of method focuses on the detection of adversarial examples. Approaches include using auxiliary neural networks~\citep{DBLP:conf/ccs/MengC17, DBLP:conf/iclr/MetzenGFB17} to decide if input is clean or adversarial, using statistical methods to verify properties of the image or the network parameters~\citep{DBLP:conf/iclr/MetzenGFB17, DBLP:conf/iclr/HendrycksG17a,DBLP:journals/corr/BhagojiCM17,DBLP:conf/iccv/LiL17}. However, Carlini and Wagner proved that under calibrated attacks these methods are still ineffective~\citep{carlini2017adversarial, carlini2017magnet}.

\section{Conclusion}

In this paper, we discovered the properties called input sensitivity that can truly discriminate $L_2$ adversarial examples from normal inputs through theoretical analysis. We further investigate the input sensitivity of $L_\infty$ attacks empirically. Based on those findings, we proposed a \textbf{G}uided \textbf{C}omplementary \textbf{D}efense module (\textbf{GCD}) integrating detection and recovery processes to verify our analysis. Our detector achieves a detection rate of over 95\% against most of the attacks. Our rectifier achieves a recovered classification accuracy of up to 99\% on MNIST, 89\% on CIFAR-10 and 87\% on ImageNet subsets. In conclusion, we are certain that at least $L_2$ adversarial examples are intrinsically different from normal inputs and shed light on using this theoretically sound property to design simple yet effective defense mechanisms.  


\bibliographystyle{bibtex/unsrtnat}
\bibliography{reference}

\begin{thebibliography}{46}
\providecommand{\natexlab}[1]{#1}
\providecommand{\url}[1]{\texttt{#1}}
\expandafter\ifx\csname urlstyle\endcsname\relax
  \providecommand{\doi}[1]{doi: #1}\else
  \providecommand{\doi}{doi: \begingroup \urlstyle{rm}\Url}\fi

\bibitem[Lin et~al.(2018)Lin, Zhang, Hsu, Skach, Haque, Tang, and
  Mars]{lin2018architectural}
Shih-Chieh Lin, Yunqi Zhang, Chang-Hong Hsu, Matt Skach, Md~E Haque, Lingjia
  Tang, and Jason Mars.
\newblock The architectural implications of autonomous driving: Constraints and
  acceleration.
\newblock In \emph{ACM SIGPLAN Notices}, volume~53, pages 751--766. ACM, 2018.

\bibitem[Parkhi et~al.(2015)Parkhi, Vedaldi, Zisserman, et~al.]{parkhi2015deep}
Omkar~M Parkhi, Andrea Vedaldi, Andrew Zisserman, et~al.
\newblock Deep face recognition.
\newblock In \emph{bmvc}, volume~1, page~6, 2015.

\bibitem[Meyes et~al.(2017)Meyes, Tercan, Roggendorf, Thiele, B{\"u}scher,
  Obdenbusch, Brecher, Jeschke, and Meisen]{meyes2017motion}
Richard Meyes, Hasan Tercan, Simon Roggendorf, Thomas Thiele, Christian
  B{\"u}scher, Markus Obdenbusch, Christian Brecher, Sabina Jeschke, and Tobias
  Meisen.
\newblock Motion planning for industrial robots using reinforcement learning.
\newblock \emph{Procedia CIRP}, 63:\penalty0 107--112, 2017.

\bibitem[Szegedy et~al.(2013)Szegedy, Zaremba, Sutskever, Bruna, Erhan,
  Goodfellow, and Fergus]{2013szegedy}
Christian Szegedy, Wojciech Zaremba, Ilya Sutskever, Joan Bruna, Dumitru Erhan,
  Ian Goodfellow, and Rob Fergus.
\newblock Intriguing properties of neural networks, 2013.

\bibitem[Akhtar and Mian(2018)]{akhtar2018threat}
Naveed Akhtar and Ajmal Mian.
\newblock Threat of adversarial attacks on deep learning in computer vision: A
  survey.
\newblock \emph{IEEE Access}, 6:\penalty0 14410--14430, 2018.

\bibitem[Madry et~al.(2018)Madry, Makelov, Schmidt, Tsipras, and
  Vladu]{madry2017pgd}
Aleksander Madry, Aleksandar Makelov, Ludwig Schmidt, Dimitris Tsipras, and
  Adrian Vladu.
\newblock Towards deep learning models resistant to adversarial attacks.
\newblock In \emph{International Conference on Learning Representations}, 2018.
\newblock URL \url{https://arxiv.org/abs/1706.06083}.

\bibitem[Goodfellow et~al.(2015)Goodfellow, Shlens, and Szegedy]{fgsm}
Ian Goodfellow, Jonathon Shlens, and Christian Szegedy.
\newblock Explaining and harnessing adversarial examples.
\newblock In \emph{International Conference on Learning Representations}, 2015.
\newblock URL \url{http://arxiv.org/abs/1412.6572}.

\bibitem[Buckman et~al.(2018)Buckman, Roy, Raffel, and
  Goodfellow]{BuckmanRRG18}
Jacob Buckman, Aurko Roy, Colin Raffel, and Ian~J. Goodfellow.
\newblock Thermometer encoding: One hot way to resist adversarial examples.
\newblock In \emph{6th International Conference on Learning Representations,
  {ICLR} 2018, Vancouver, BC, Canada, April 30 - May 3, 2018, Conference Track
  Proceedings}, 2018.
\newblock URL \url{https://openreview.net/forum?id=S18Su--CW}.

\bibitem[Dhillon et~al.(2018)Dhillon, Azizzadenesheli, Lipton, Bernstein,
  Kossaifi, Khanna, and Anandkumar]{DhillonALBKKA18}
Guneet~S. Dhillon, Kamyar Azizzadenesheli, Zachary~C. Lipton, Jeremy Bernstein,
  Jean Kossaifi, Aran Khanna, and Animashree Anandkumar.
\newblock Stochastic activation pruning for robust adversarial defense.
\newblock In \emph{6th International Conference on Learning Representations,
  {ICLR} 2018, Vancouver, BC, Canada, April 30 - May 3, 2018, Conference Track
  Proceedings}, 2018.
\newblock URL \url{https://openreview.net/forum?id=H1uR4GZRZ}.

\bibitem[Samangouei et~al.(2018)Samangouei, Kabkab, and
  Chellappa]{SamangoueiKC18}
Pouya Samangouei, Maya Kabkab, and Rama Chellappa.
\newblock Defense-gan: Protecting classifiers against adversarial attacks using
  generative models.
\newblock In \emph{6th International Conference on Learning Representations,
  {ICLR} 2018, Vancouver, BC, Canada, April 30 - May 3, 2018, Conference Track
  Proceedings}, 2018.
\newblock URL \url{https://openreview.net/forum?id=BkJ3ibb0-}.

\bibitem[Xie et~al.(2018)Xie, Wang, Zhang, Ren, and
  Yuille]{DBLP:conf/iclr/XieWZRY18}
Cihang Xie, Jianyu Wang, Zhishuai Zhang, Zhou Ren, and Alan~L. Yuille.
\newblock Mitigating adversarial effects through randomization.
\newblock In \emph{6th International Conference on Learning Representations,
  {ICLR} 2018, Vancouver, BC, Canada, April 30 - May 3, 2018, Conference Track
  Proceedings}, 2018.
\newblock URL \url{https://openreview.net/forum?id=Sk9yuql0Z}.

\bibitem[Song et~al.(2018{\natexlab{a}})Song, Kim, Nowozin, Ermon, and
  Kushman]{DBLP:conf/iclr/SongKNEK18}
Yang Song, Taesup Kim, Sebastian Nowozin, Stefano Ermon, and Nate Kushman.
\newblock Pixeldefend: Leveraging generative models to understand and defend
  against adversarial examples.
\newblock In \emph{6th International Conference on Learning Representations,
  {ICLR} 2018, Vancouver, BC, Canada, April 30 - May 3, 2018, Conference Track
  Proceedings}, 2018{\natexlab{a}}.
\newblock URL \url{https://openreview.net/forum?id=rJUYGxbCW}.

\bibitem[Guo et~al.(2018{\natexlab{a}})Guo, Rana, Ciss{\'{e}}, and van~der
  Maaten]{GuoRCM18}
Chuan Guo, Mayank Rana, Moustapha Ciss{\'{e}}, and Laurens van~der Maaten.
\newblock Countering adversarial images using input transformations.
\newblock In \emph{6th International Conference on Learning Representations,
  {ICLR} 2018, Vancouver, BC, Canada, April 30 - May 3, 2018, Conference Track
  Proceedings}, 2018{\natexlab{a}}.
\newblock URL \url{https://openreview.net/forum?id=SyJ7ClWCb}.

\bibitem[Meng and Chen(2017)]{DBLP:conf/ccs/MengC17}
Dongyu Meng and Hao Chen.
\newblock Magnet: {A} two-pronged defense against adversarial examples.
\newblock In \emph{Proceedings of the 2017 {ACM} {SIGSAC} Conference on
  Computer and Communications Security, {CCS} 2017, Dallas, TX, USA, October 30
  - November 03, 2017}, pages 135--147, 2017.
\newblock \doi{10.1145/3133956.3134057}.
\newblock URL \url{https://doi.org/10.1145/3133956.3134057}.

\bibitem[Metzen et~al.(2017)Metzen, Genewein, Fischer, and
  Bischoff]{DBLP:conf/iclr/MetzenGFB17}
Jan~Hendrik Metzen, Tim Genewein, Volker Fischer, and Bastian Bischoff.
\newblock On detecting adversarial perturbations.
\newblock In \emph{5th International Conference on Learning Representations,
  {ICLR} 2017, Toulon, France, April 24-26, 2017, Conference Track
  Proceedings}, 2017.
\newblock URL \url{https://openreview.net/forum?id=SJzCSf9xg}.

\bibitem[Hendrycks and Gimpel(2017)]{DBLP:conf/iclr/HendrycksG17a}
Dan Hendrycks and Kevin Gimpel.
\newblock Early methods for detecting adversarial images.
\newblock In \emph{5th International Conference on Learning Representations,
  {ICLR} 2017, Toulon, France, April 24-26, 2017, Workshop Track Proceedings},
  2017.
\newblock URL \url{https://openreview.net/forum?id=B1dexpDug}.

\bibitem[Bhagoji et~al.(2017)Bhagoji, Cullina, and
  Mittal]{DBLP:journals/corr/BhagojiCM17}
Arjun~Nitin Bhagoji, Daniel Cullina, and Prateek Mittal.
\newblock Dimensionality reduction as a defense against evasion attacks on
  machine learning classifiers.
\newblock \emph{CoRR}, abs/1704.02654, 2017.
\newblock URL \url{http://arxiv.org/abs/1704.02654}.

\bibitem[Li and Li(2017)]{DBLP:conf/iccv/LiL17}
Xin Li and Fuxin Li.
\newblock Adversarial examples detection in deep networks with convolutional
  filter statistics.
\newblock In \emph{{IEEE} International Conference on Computer Vision, {ICCV}
  2017, Venice, Italy, October 22-29, 2017}, pages 5775--5783, 2017.
\newblock \doi{10.1109/ICCV.2017.615}.
\newblock URL \url{https://doi.org/10.1109/ICCV.2017.615}.

\bibitem[Carlini and Wagner(2017{\natexlab{a}})]{carlini2017adversarial}
Nicholas Carlini and David Wagner.
\newblock Adversarial examples are not easily detected: Bypassing ten detection
  methods.
\newblock In \emph{Proceedings of the 10th ACM Workshop on Artificial
  Intelligence and Security}, pages 3--14. ACM, 2017{\natexlab{a}}.

\bibitem[Dhaliwal and Shintre(2018)]{dhaliwal2018gradient}
Jasjeet Dhaliwal and Saurabh Shintre.
\newblock Gradient similarity: An explainable approach to detect adversarial
  attacks against deep learning.
\newblock \emph{arXiv preprint arXiv:1806.10707}, 2018.

\bibitem[Zhang et~al.(2018)Zhang, Yang, and Ye]{2018qinghua_imagegrad}
Chiliang Zhang, Zhimou Yang, and Zuochang Ye.
\newblock Detecting adversarial perturbations with salieny, 2018.
\newblock URL \url{http://dx.doi.org/10.1145/3301551.3301588}.

\bibitem[Moosavi-Dezfooli et~al.(2016)Moosavi-Dezfooli, Fawzi, and
  Frossard]{moosavi2016deepfool}
Seyed-Mohsen Moosavi-Dezfooli, Alhussein Fawzi, and Pascal Frossard.
\newblock Deepfool: a simple and accurate method to fool deep neural networks.
\newblock In \emph{Proceedings of the IEEE conference on computer vision and
  pattern recognition}, pages 2574--2582, 2016.

\bibitem[Carlini and Wagner(2017{\natexlab{b}})]{carlini2017cw}
Nicholas Carlini and David Wagner.
\newblock Towards evaluating the robustness of neural networks.
\newblock In \emph{2017 IEEE Symposium on Security and Privacy (SP)}, pages
  39--57. IEEE, 2017{\natexlab{b}}.

\bibitem[Rony et~al.(2019)Rony, Hafemann, Oliveira, Ayed, Sabourin, and
  Granger]{rony2019ddn}
J{\'e}r{\^o}me Rony, Luiz~G Hafemann, Luiz~S Oliveira, Ismail~Ben Ayed, Robert
  Sabourin, and Eric Granger.
\newblock Decoupling direction and norm for efficient gradient-based l2
  adversarial attacks and defenses.
\newblock In \emph{Proceedings of the IEEE Conference on Computer Vision and
  Pattern Recognition}, pages 4322--4330, 2019.

\bibitem[Yu et~al.(2019)Yu, Hu, Guo, Chao, and
  Weinberger]{Yu_2019_NeurIPSweakness}
Tao Yu, Shengyuan Hu, Chuan Guo, Weilun Chao, and Kilian Weinberger.
\newblock A new defense against adversarial images: Turning a weakness into a
  strength.
\newblock In \emph{Proceedings of the 33rd Conference on Neural Information
  Processing Systems (NeurIPS 2019)}, Oct. 2019.

\bibitem[Gong et~al.(2017)Gong, Wang, and Ku]{DBLP:journals/corr/GongWK17}
Zhitao Gong, Wenlu Wang, and Wei{-}Shinn Ku.
\newblock Adversarial and clean data are not twins.
\newblock \emph{CoRR}, abs/1704.04960, 2017.
\newblock URL \url{http://arxiv.org/abs/1704.04960}.

\bibitem[LeCun et~al.(1998)LeCun, Bottou, Bengio, Haffner,
  et~al.]{lecun1998mnist}
Yann LeCun, L{\'e}on Bottou, Yoshua Bengio, Patrick Haffner, et~al.
\newblock Gradient-based learning applied to document recognition.
\newblock \emph{Proceedings of the IEEE}, 86\penalty0 (11):\penalty0
  2278--2324, 1998.

\bibitem[Krizhevsky et~al.(2009)]{krizhevsky2009cifar}
Alex Krizhevsky et~al.
\newblock Learning multiple layers of features from tiny images.
\newblock Technical report, Citeseer, 2009.

\bibitem[Deng et~al.(2009)Deng, Dong, Socher, Li, Li, and
  Fei-Fei]{imagenet_cvpr09}
J.~Deng, W.~Dong, R.~Socher, L.-J. Li, K.~Li, and L.~Fei-Fei.
\newblock {ImageNet: A Large-Scale Hierarchical Image Database}.
\newblock In \emph{CVPR09}, 2009.

\bibitem[Simonyan and Zisserman(2015)]{vgg11}
Karen Simonyan and Andrew Zisserman.
\newblock Very deep convolutional networks for large-scale image recognition.
\newblock In \emph{3rd International Conference on Learning Representations,
  {ICLR} 2015, San Diego, CA, USA, May 7-9, 2015, Conference Track
  Proceedings}, 2015.
\newblock URL \url{http://arxiv.org/abs/1409.1556}.

\bibitem[He et~al.(2016)He, Zhang, Ren, and Sun]{resnet}
Kaiming He, Xiangyu Zhang, Shaoqing Ren, and Jian Sun.
\newblock Deep residual learning for image recognition.
\newblock In \emph{2016 {IEEE} Conference on Computer Vision and Pattern
  Recognition, {CVPR} 2016, Las Vegas, NV, USA, June 27-30, 2016}, pages
  770--778, 2016.
\newblock \doi{10.1109/CVPR.2016.90}.
\newblock URL \url{https://doi.org/10.1109/CVPR.2016.90}.

\bibitem[Hu et~al.(2019)Hu, Yu, Guo, Chao, and Weinberger]{tws}
Shengyuan Hu, Tao Yu, Chuan Guo, Wei{-}Lun Chao, and Kilian~Q. Weinberger.
\newblock A new defense against adversarial images: Turning a weakness into a
  strength.
\newblock In \emph{Advances in Neural Information Processing Systems 32: Annual
  Conference on Neural Information Processing Systems 2019, NeurIPS 2019, 8-14
  December 2019, Vancouver, BC, Canada}, pages 1633--1644, 2019.

\bibitem[Song et~al.(2018{\natexlab{b}})Song, Kim, Nowozin, Ermon, and
  Kushman]{pixeldefend}
Yang Song, Taesup Kim, Sebastian Nowozin, Stefano Ermon, and Nate Kushman.
\newblock Pixeldefend: Leveraging generative models to understand and defend
  against adversarial examples.
\newblock In \emph{6th International Conference on Learning Representations,
  {ICLR} 2018, Vancouver, BC, Canada, April 30 - May 3, 2018, Conference Track
  Proceedings}, 2018{\natexlab{b}}.
\newblock URL \url{https://openreview.net/forum?id=rJUYGxbCW}.

\bibitem[Lee et~al.(2018)Lee, Lee, Lee, and Shin]{mds}
Kimin Lee, Kibok Lee, Honglak Lee, and Jinwoo Shin.
\newblock A simple unified framework for detecting out-of-distribution samples
  and adversarial attacks.
\newblock In \emph{Advances in Neural Information Processing Systems 31: Annual
  Conference on Neural Information Processing Systems 2018, NeurIPS 2018, 3-8
  December 2018, Montr{\'{e}}al, Canada}, pages 7167--7177, 2018.

\bibitem[Guo et~al.(2018{\natexlab{b}})Guo, Rana, Ciss{\'{e}}, and van~der
  Maaten]{tvm}
Chuan Guo, Mayank Rana, Moustapha Ciss{\'{e}}, and Laurens van~der Maaten.
\newblock Countering adversarial images using input transformations.
\newblock In \emph{6th International Conference on Learning Representations,
  {ICLR} 2018, Vancouver, BC, Canada, April 30 - May 3, 2018, Conference Track
  Proceedings}, 2018{\natexlab{b}}.
\newblock URL \url{https://openreview.net/forum?id=SyJ7ClWCb}.

\bibitem[Kurakin et~al.(2017{\natexlab{a}})Kurakin, Goodfellow, and
  Bengio]{bim}
Alexey Kurakin, Ian Goodfellow, and Samy Bengio.
\newblock Adversarial examples in the physical world.
\newblock \emph{International Conference on Learning Representations Workshop},
  2017{\natexlab{a}}.
\newblock URL \url{https://arxiv.org/abs/1607.02533}.

\bibitem[Su et~al.(2019)Su, Vargas, and Sakurai]{su2019onepixel}
Jiawei Su, Danilo~Vasconcellos Vargas, and Kouichi Sakurai.
\newblock One pixel attack for fooling deep neural networks.
\newblock \emph{IEEE Transactions on Evolutionary Computation}, 2019.

\bibitem[Papernot et~al.(2016{\natexlab{a}})Papernot, McDaniel, and
  Goodfellow]{papernot2016transferability}
Nicolas Papernot, Patrick McDaniel, and Ian Goodfellow.
\newblock Transferability in machine learning: from phenomena to black-box
  attacks using adversarial samples.
\newblock \emph{arXiv preprint arXiv:1605.07277}, 2016{\natexlab{a}}.

\bibitem[Papernot et~al.(2017)Papernot, McDaniel, Goodfellow, Jha, Celik, and
  Swami]{papernot2017practical}
Nicolas Papernot, Patrick McDaniel, Ian Goodfellow, Somesh Jha, Z~Berkay Celik,
  and Ananthram Swami.
\newblock Practical black-box attacks against machine learning.
\newblock In \emph{Proceedings of the 2017 ACM on Asia conference on computer
  and communications security}, pages 506--519. ACM, 2017.

\bibitem[Alzantot et~al.(2019)Alzantot, Sharma, Chakraborty, Zhang, Hsieh, and
  Srivastava]{alzantot2019genattack}
Moustafa Alzantot, Yash Sharma, Supriyo Chakraborty, Huan Zhang, Cho-Jui Hsieh,
  and Mani~B Srivastava.
\newblock Genattack: Practical black-box attacks with gradient-free
  optimization.
\newblock In \emph{Proceedings of the Genetic and Evolutionary Computation
  Conference}, pages 1111--1119. ACM, 2019.

\bibitem[Guo et~al.(2019)Guo, Gardner, You, Wilson, and
  Weinberger]{pmlr-v97-guo19a}
Chuan Guo, Jacob Gardner, Yurong You, Andrew~Gordon Wilson, and Kilian
  Weinberger.
\newblock Simple black-box adversarial attacks.
\newblock In Kamalika Chaudhuri and Ruslan Salakhutdinov, editors,
  \emph{Proceedings of the 36th International Conference on Machine Learning},
  volume~97 of \emph{Proceedings of Machine Learning Research}, pages
  2484--2493, Long Beach, California, USA, 09--15 Jun 2019. PMLR.
\newblock URL \url{http://proceedings.mlr.press/v97/guo19a.html}.

\bibitem[Kurakin et~al.(2017{\natexlab{b}})Kurakin, Goodfellow, and
  Bengio]{scale17}
Alexey Kurakin, Ian~J. Goodfellow, and Samy Bengio.
\newblock Adversarial machine learning at scale.
\newblock In \emph{5th International Conference on Learning Representations,
  {ICLR} 2017, Toulon, France, April 24-26, 2017, Conference Track
  Proceedings}, 2017{\natexlab{b}}.
\newblock URL \url{https://openreview.net/forum?id=BJm4T4Kgx}.

\bibitem[Papernot et~al.(2016{\natexlab{b}})Papernot, McDaniel, Wu, Jha, and
  Swami]{DBLP:conf/sp/PapernotM0JS16}
Nicolas Papernot, Patrick~D. McDaniel, Xi~Wu, Somesh Jha, and Ananthram Swami.
\newblock Distillation as a defense to adversarial perturbations against deep
  neural networks.
\newblock In \emph{{IEEE} Symposium on Security and Privacy, {SP} 2016, San
  Jose, CA, USA, May 22-26, 2016}, pages 582--597, 2016{\natexlab{b}}.
\newblock \doi{10.1109/SP.2016.41}.
\newblock URL \url{https://doi.org/10.1109/SP.2016.41}.

\bibitem[Athalye et~al.(2018)Athalye, Carlini, and
  Wagner]{DBLP:conf/icml/AthalyeC018}
Anish Athalye, Nicholas Carlini, and David~A. Wagner.
\newblock Obfuscated gradients give a false sense of security: Circumventing
  defenses to adversarial examples.
\newblock In \emph{Proceedings of the 35th International Conference on Machine
  Learning, {ICML} 2018, Stockholmsm{\"{a}}ssan, Stockholm, Sweden, July 10-15,
  2018}, pages 274--283, 2018.
\newblock URL \url{http://proceedings.mlr.press/v80/athalye18a.html}.

\bibitem[Carlini and Wagner(2016)]{DBLP:journals/corr/CarliniW16}
Nicholas Carlini and David~A. Wagner.
\newblock Defensive distillation is not robust to adversarial examples.
\newblock \emph{CoRR}, abs/1607.04311, 2016.
\newblock URL \url{http://arxiv.org/abs/1607.04311}.

\bibitem[Carlini and Wagner(2017{\natexlab{c}})]{carlini2017magnet}
Nicholas Carlini and David Wagner.
\newblock Magnet and" efficient defenses against adversarial attacks" are not
  robust to adversarial examples.
\newblock \emph{arXiv preprint arXiv:1711.08478}, 2017{\natexlab{c}}.

\end{thebibliography}

\end{document}